# ARTIFICIAL NEOTENY IN EVOLUTIONARY IMAGE SEGMENTATION

**Vitorino Ramos**
CVRM-IST Geo-Systems Centre,
Instituto Superior Técnico
Av. Rovisco Pais, 1049-001, Lisboa, Portugal
vitorino.ramos@alfa.ist.utl.pt

**Abstract**

*Neoteny,* also spelled *Paedomorphosis*, can be defined in biological terms as the retention by an organism of juvenile or even larval traits into later life. In some species, all morphological development is retarded; the organism is juvenilized but sexually mature. Such shifts of reproductive capability would appear to have adaptive significance to organisms that exhibit it. In terms of evolutionary theory, the process of *paedomorphosis* suggests that larval stages and developmental phases of existing organisms may give rise, under certain circumstances, to wholly new organisms. Although the present work does not pretend to model or simulate the biological details of such a concept in any way, these ideas were incorporated by a rather simple abstract computational strategy, in order to allow (if possible) for faster convergence into simple non-memetic Genetic Algorithms, i.e. without using local improvement procedures (e.g. via *Baldwin* or *Lamarckian* learning). As a case-study, the Genetic Algorithm was used for colour image segmentation purposes by using *K*-mean unsupervised clustering methods, namely for guiding the evolutionary algorithm in his search for finding the optimal or sub-optimal data partition. Average results suggest that the use of *neotonic* strategies by employing juvenile genotypes into the later generations and the use of linear-dynamic mutation rates instead of constant, can increase fitness values by 58% comparing to classical Genetic Algorithms, independently from the starting population characteristics on the search space.

## 1. INCORPORATING NEOTENY INTO GENETIC ALGORITHMS

Evolution is carried out by a process dependent on mutation and natural selection. Expositions of this thesis, however, tend to overlook the fact that mutation occurs in the genotype, whereas natural selection acts only on the phenotype, the organism produced. It follows from this that the theory of evolution requires as one of its essential parts a consideration of the developmental or epigenetic processes (due to external not genetic influences) by which the genotype becomes translated into the phenotype. Natural selection as visualised by *Darwin*, results in the production by one generation of offspring that are able to survive and reproduce themselves to form a further generation. The time unit appropriate to natural selection is therefore the generation interval. There will always be some natural selective pressure for the shortening of the generation interval, simply out of a natural economy, and for an increase of the number of offspring produced by any reproducing individual. One of the ways in which such an increase could be assured would be the lengthening of the reproductive phase in the life history; another would be an increase in the number of offspring produced [6,7,21].

There are, of course, not only natural selective pressures that operate. It is clear enough that, in evolution, they have often been overcome by other pressures. There is another natural selective of more general importance. This is the pressure to restrict the length of the reproductive period, and indeed to remove reproductive individuals, in order to make room for the maturation of a new generation in which new genetic combinations can be tried out for their fitness. A species whose individuals were immortal would exhaust its possibilities for future evolution as soon as its numbers saturated all the ecological niches suitable for its way of life. Death is a necessary condition for the trying out of new genetic combinations in later generations. It is usually brought about, in great part at least, by combinations of two processes: restriction of the period of effective reproduction to a certain portion of a life history, and as a necessary condition of this, the absence of natural selection for genetic mutations that would be effective in preserving life after reproduction has ceased. Such types of development offer the possibility of changing the relative importance of various stages in relation to the exploitation of resources and reproduction by the species. There are, for instance, many types of animals (particularly insects; e.g. termites) in which nearly the whole life history is passed in a larval stage in which most of the feeding and growth of the organism is carried out, the final adult stage being short and used almost entirely for reproduction (in simple terms, they can be seen as genetic material *carriers*). Another evolutionary strategy has been to transfer the reproductive phase from the final stage of the life history to some earlier larval stage. This again has occurred in certain insects. If such a process is carried to its logical evolutionary conclusion, the final previously adult stage of the life history may totally disappear, the larval stage of the earlier evolutionary form becoming the adult stage of the later derivative of it. It has been suggested that such process of *Neoteny* (i.e., the retention of juvenile characteristics in adulthood - the term was coined by *Kollman*) have played a decision role in certain earlier phases of evolution, evidence of which is now lost. It has been argued also, that the whole vertebrate *phylum* may have originated from modifications of one of the larval stages of an invertebrate group [21].

As his known, the process of diversity loss in genetic algorithms is often the cause of premature convergence, and as a consequence, the early convergence to an inferior local maximum. A large number of existing techniques are used to maintain diversity in *Genetic Algorithms* (GA, [4,10,16]). These include maintaining large population sizes, employing low reproductive or parent-selection pressures, applying mutation to the genotype, restarting the GA with new random genotypes, employing parallel populations (with occasional interchanging of fit chromosomes between populations), and niche-formation techniques. The present approach uses the concept of *Neoteny*. This last strategy was then incorporated in simple non-memetic genetic algorithms, by simply preserving some individuals in the earlier generations (using elitism), and by randomly re-injecting this genetic material into the later generations, allowing for substantial increases in diversity, and (as it seems) for an appropriate balance between exploration and exploitation of the search space. Some questions, however, should be discussed. For instance, at which period in the whole evolutionary process should this *neotonic* individuals be captured, how many should be thrown in (in the later generations), and when thus this *throwing* process should start? Section 3 is dedicated to those questions. Finally, and in order to study the impact of such abstract concept, yet computationally possible by using sequential incorporations of older genetic material, a difficult combinatorial problem was chosen: colour image segmentation. The present work is divided into testing and discussing dynamic mutation rates, the implementation and testing of artificial *Neoteny* Strategies, and finally the respective conclusions are presented.

## 2. TESTING AND DISCUSSING DYNAMIC MUTATION RATES

Since the search space can be huge for similar applications (consider for example, satellite images or normal images at higher resolutions), and in order to speed-up the GA convergence (if possible), some experiments were conducted with dynamic mutation rates (i.e. time-dependent). As pointed by *Rudolph* [19] in 1994, one possible route to achieve global optimal convergence might be the introduction of time varying mutation *and* selection probabilities. *Rudolph* suggests to use two simultaneous strategies instead of one, referring the work of *Davis* (1991, [5]), where it has been shown that the introduction of time varying mutation probabilities alone does not help. Anyway, all experiments were conducted in one-point crossover genetic algorithms ($p_c$ = constant = 0.8), with 100 individuals (each pair of individuals selected via roulette wheel selection and windowing scaling, yields two new individuals), and within 3000 generations. Each individual was represented by a binary vector of length $n$ = 531 (each 3 bit can code up to 8 colour clusters, although only 6 are needed, since only 6 prominent colours are present in this maps / 177 colour small cubes present). In these conditions, each generation $g$ takes on average 0.0693 seconds (PENTIUM II - 333MHz / 128Mb RAM), which means about 3.5 minutes on 3000 generations (except for test #9, $g_{max}$ = 6000 - see table 1 / image with $500^2$ pixels and 214385 different colours). Then, 2 tests were run with constant mutation rates $p_m$ = 0.15 (table 1 / column D=*C*), 8 with linear-dynamic mutation rates (table 1 / column D=*LD*), and finally 25 with quadratic-dynamic mutation rates (table 1 / column D=*QD*). *C*, *LD* and *QD* tests can be expressed by the following mutation rate expressions:

- $C \Rightarrow p_m = 0.15; g \in [0, 3000]$
- $LD \Rightarrow p_m = 0.15\ (g=0)\ /\ p_m = 0.15/g\ ; g \in [1, 100]\ /\ p_m = 0.0015\ ; g \in [101, 3000]$
- $QD \Rightarrow p_m = 0.15\ (g=0)\ /\ p_m = 0.15/g^2; g \in [1, 100]\ /\ p_m = 0.000015; g \in [101, 3000]$

Many other functions were tried, some of them inspired on *Simulated Annealing* methods (SA, [15]) or in variants of it (e.g. *Adaptive Simulated Annealing*, *Re-Annealing*, *Quenching*, [12,13,14]), as the present problem seems similar [5]. In fact, both methods are applied in search-combinatorial-optimization problems, and both start from random points in the search space. Particularly interesting in the present case is that, the mutation rate in GA[s] can be seen as the temperature parameter in SA[s] (they both affect the convergence of the respective strategy and the balance between an appropriate exploring/exploiting character of the algorithm). Similarly, scheduling temperatures in SA[s] (one of the most difficult problems to solve for this method) can be seen as the implementation of dynamic mutations on GA[s]. Surprisingly (and even if several SA temperature scheduling rates were tried, generally of logarithmic or exponential nature [11,12,13,14,18,22]), the GA mutation settings that yields the best results were always the simplest ones (i.e. *LD* and *QD* - see table 2 for average results). Another fact, seems to be that the best dynamic rate should change with the starting population (compare for instance tests #2,11 and #3,12), suggesting that possibly the optimal mutation probability depends on the search landscape, the GA coding (introducing or not a multi-optimisation problem and eventually several genotype mappings to the same phenotype), and finally on the objective function itself. All the previous results appear to be in some accordance with those from *Bäck* [1,2,3] and *Mühlenbein* [17] (namely, in the hyperbolic nature of the functions used). Independently of each other, the two authors investigated in 1992, the optimal mutation rate for a simple (1+1)-algorithm (a single parent generates an offspring by means of mutation and the better of both survives for the next generation) with the objective function $f(x) = \sum_{i=1}^{n} x_I$ ("counting ones"). As putted by *Bäck* [3] the optimal mutation probability depends strongly on the objective function value $f(x)$ and follows a hyperbolic law of the form $p_m = (2.(f(x)+1)-n)^{-1}$. In order to model the hyperbolic shape of the last equation, independently of the objective function, *Bäck* used a time-dependent mutation rate $p_m(g)$ (where $n$ denotes the chromosome length, and $T$ a given maximum of generations $g$). From the condition $p_m(T-1)=1/n$, the hyperbolic formulation $p_m = (a+b.g)^{-1}$ then yields (Eq.1):

$$p_m(g) = \left( p_m^{-1}(0) + \frac{n - p_m^{-1}(0)}{T-1} \cdot g \right)^{-1} \qquad (1)$$

| Test # | A | B | C | D | E | F | G | H |
|---|---|---|---|---|---|---|---|---|
| 1 | 9 | 3000 | 0.8 | C | 0 | - | - | 201.611623 |
| **2** | **9** | **3000** | **0.8** | **LD** | **0** | **-** | **-** | **325.528410** |
| 3 | 9 | 3000 | 0.8 | QD | 0 | - | - | 312.694066 |
| 4 | 9 | 3000 | 0.8 | B [0.15] | 0 | - | - | 203.964332 |
| 5 | 9 | 3000 | 0.8 | B [0.50] | 0 | - | - | 180.236736 |
| **6** | **9** | **3000** | **0.8** | **LD** | **1** | **[1,100]** | **[1000,3000]** | **326.426236** |
| 7 | 9 | 3000 | 0.8 | QD | 1 | [1,100] | [1000,3000] | 314.125107 |
| 8 | 9 | 3000 | 0.8 | B [0.15] | 1 | [1,100] | [1000,3000] | 207.823020 |
| **9** | **9** | **6000** | **0.8** | **LD** | **0** | **-** | **-** | **326.993288** |
| 10 | 7445 | 3000 | 0.8 | C | 0 | - | - | 191.146788 |
| 11 | 7445 | 3000 | 0.8 | LD | 0 | - | - | 306.475341 |
| **12** | **7445** | **3000** | **0.8** | **QD** | **0** | **-** | **-** | **326.549272** |
| 13 | 7445 | 3000 | 0.8 | LD | 1 | [1,100] | [1000,3000] | 308.919431 |
| 14 | 7445 | 3000 | 0.8 | QD | 1 | [1,100] | [1000,3000] | 321.010773 |
| 15 | 7445 | 3000 | 0.8 | QD | 1 | [1,100] | [500,3000] | 319.063587 |
| 16 | 7445 | 3000 | 0.8 | QD | 1 | [1,100] | [350,3000] | 320.481312 |
| 17 | 7445 | 3000 | 0.8 | QD | 1 | [1,100] | [320,3000] | 316.784335 |
| 18 | 7445 | 3000 | 0.8 | QD | 1 | [1,100] | [300,3000] | 322.772565 |
| 19 | 7445 | 3000 | 0.8 | QD | 1 | [1,100] | [285,3000] | 316.366818 |
| **20** | **7445** | **3000** | **0.8** | **QD** | **1** | **[1,100]** | **[280,3000]** | **324.908299** |
| 21 | 7445 | 3000 | 0.8 | QD | 1 | [1,100] | [279,3000] | 317.947635 |
| 22 | 7445 | 3000 | 0.8 | QD | 1 | [1,100] | [277,3000] | 318.843974 |
| 23 | 7445 | 3000 | 0.8 | QD | 1 | [1,100] | [275,3000] | 319.100290 |
| 24 | 7445 | 3000 | 0.8 | QD | 1 | [1,100] | [200,3000] | 316.244083 |
| 25 | 7445 | 3000 | 0.8 | QD | 1 | [1,100] | [150,3000] | 316.556148 |
| 26 | 9 | 3000 | 0.8 | LD | 2 | [1,100] | [1000,3000] | 319.990759 |
| 27 | 7445 | 3000 | 0.8 | QD | 1.5 | [1,100] | [280,3000] | 312.452034 |
| 28 | 7445 | 3000 | 0.8 | QD | 2 | [1,100] | [280,3000] | 311.933670 |
| 29 | 7445 | 3000 | 0.8 | QD | 3 | [1,100] | [280,3000] | 303.136676 |
| 30 | 7445 | 3000 | 0.8 | QD | 5 | [1,100] | [280,3000] | 297.281200 |
| 31 | 7445 | 3000 | 0.8 | QD | 1 | [100,200] | [1000,3000] | 317.683124 |
| 32 | 7445 | 3000 | 0.8 | QD | 1 | [100,200] | [280,3000] | 309.894177 |
| 33 | 7445 | 3000 | 0.8 | QD | 1 | [1,50] | [280,3000] | 322.543241 |
| 34 | 7445 | 3000 | 0.8 | QD | 1 | [1,30] | [280,3000] | 317.450920 |
| 35 | 9 | 3000 | 0.8 | LD | 2* | [1,100] | [1000,3000] | 323.254605 |
| 36 | 9 | 3000 | 0.8 | QD | 2* | [1,100] | [1000,3000] | 315.927842 |
| 37 | 7445 | 3000 | 0.8 | LD | 2* | [1,100] | [1000,3000] | 290.810651 |
| **38** | **7445** | **3000** | **0.8** | **QD** | **2*** | **[1,100]** | **[1000,3000]** | **326.281866** |

Table 1 - Results for 38 GA runs in terms of fitness (column H: $10^9/J$). Column A: Random seed; Column B: maximum number of generations; Column C: Crossover probability; Column D: Type of Mutation (*C*=constant=0.15, *LD* or *QD* / B = *Bäck*'s function with $p_m(0)= ½$ or $p_m(0)=0.15$); Column E: average number of *Neotonic* individuals re-injected in the generation interval at column G (*one individual completely random created re-injected with one *Neotonic* individual); Column F: Generation interval where *Neotonic* individuals were captured (one for each generation).

| GA Strategy (see Test # - Table 1) | | R=9 | R=7445 | R=917 | R=14 | R=27 | Average |
|---|---|---|---|---|---|---|---|
| C | (1,7) | 201.61 | 191.15 | 183.36 | 205.07 | 201.52 | 196.54 |
| LD | (2,8) | 325.53 | 306.48 | 275.41 | 322.07 | **314.29** | 308.76 |
| QD | (3,9) | 312.69 | **326.55** | 286.61 | 290.89 | 270.74 | 297.50 |
| LD/N | (4,10) | **326.43** | 308.92 | 285.14 | **323.07** | 313.87 | **311.49** |
| QD/N | (5,11) | 314.13 | 321.01 | **310.42** | 281.46 | 279.56 | 301.32 |
| LD/N+R | (35,37) | 323.25 | 290.81 | 284.94 | 317.57 | 312.83 | 305.88 |
| QD/N+R | (36,38) | 315.93 | 326.28 | 292.87 | 297.81 | 305.70 | 307.72 |
| Average | | 302.80 | 295.89 | 274.11 | 291.13 | 285.50 | 289.89 |

Table 2 - Analysis of different GA strategies with different starting populations (see Table 1 for similar test types / values for 3000 generations / best values for each random seed are in bold).

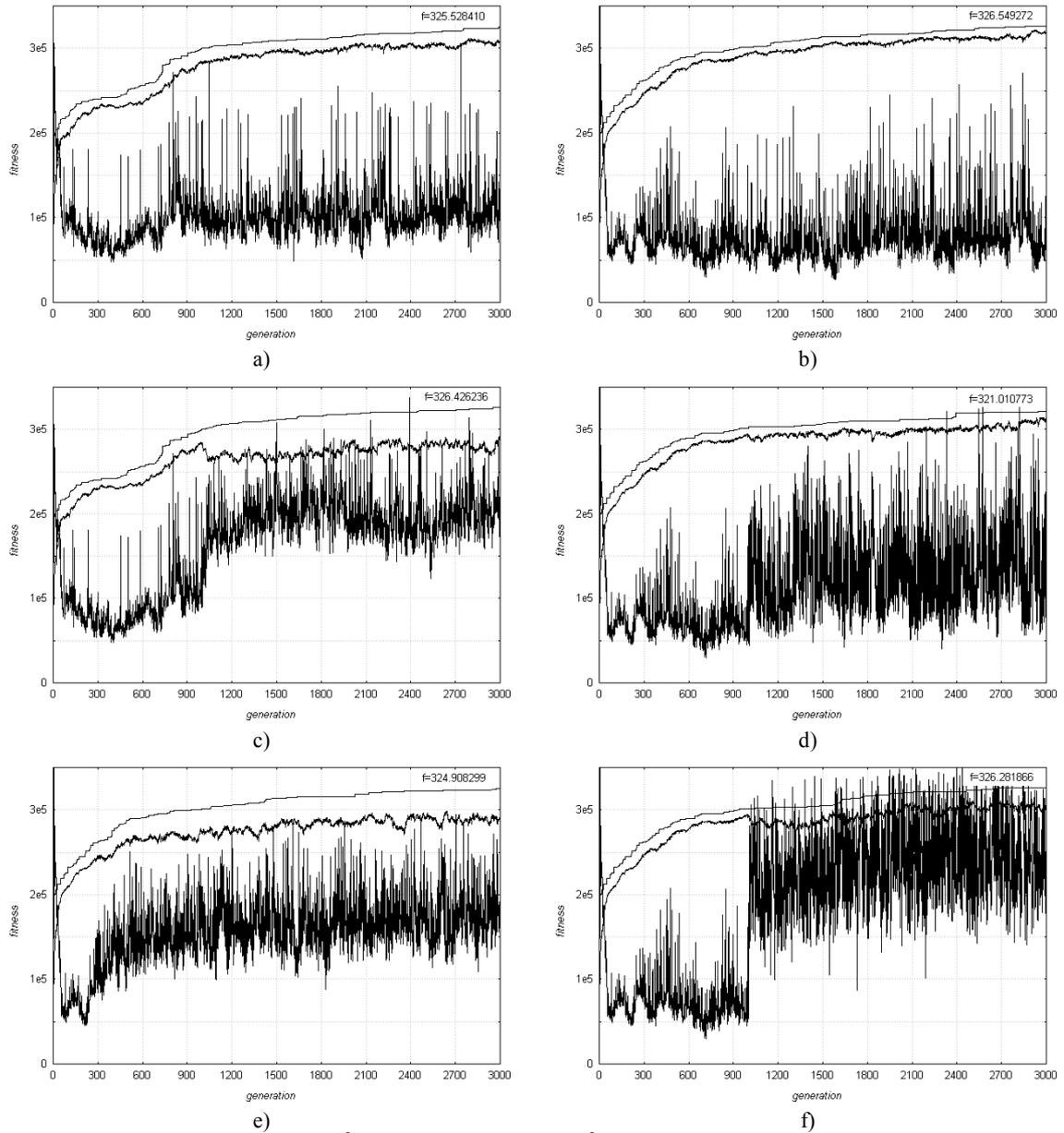

Fig. 2 - Best individual fitness ($10^9/J$), average fitness ($10^9/J$) and fitness standard deviation (x10) for each generation (100 individuals). a) *LD*, *R*=9 (test #2 - see Table 1). b) *QD*, *R*=7445 (test #12). c) *LD/N*, *R*=9 (test #6). d) *QD/N*, *R*=7445 (test #14). e) *QD/N*, *R*=7445, and incorporating *Neotonic* individuals for *g*>280 (test #20). f) *QD/N+R*, *R*=7445 (test #38).

| Random Seed | Best Fitness | Worst Fitness | Average Fitness | Std.Dev. | Sum |
|---|---|---|---|---|---|
| R=9 | 126.1 | 81.0 | 104.8 | 12.2 | 10484.0 |
| R=14 | 125.6 | 78.4 | 100.7 | 11.8 | 10068.7 |
| R=27 | 128.5 | 78.2 | 100.1 | 11.7 | 10005.4 |
| R=917 | 132.9 | 79.0 | 101.3 | 12.3 | 10125.8 |
| R=7445 | 128.5 | 79.3 | 101.9 | 12.1 | 10192.1 |

Table 3 – Random seed effect on the initial population, in terms of fitness ($10^9/J$) for *g*=0 (100 chromosomes)

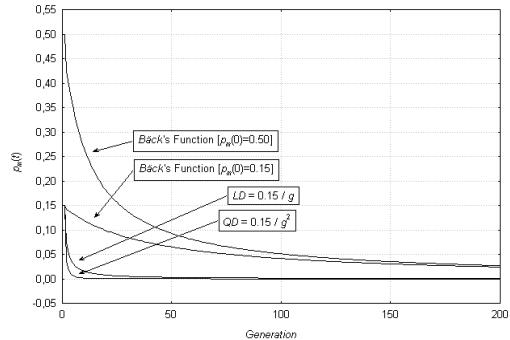

Fig. 3 - Comparing *Bäck*'s [1,3] dynamical mutation rate function (with $p_m(0)= ½$ and $p_m(0)=0.15$) and *LD*, *QD* functions, for $g \in [0,200]$.

There is however, at least one substantial difference. As mentioned by *Bäck* based on his own research and on *Muhlenbein*'s work, practical applications of genetic algorithms often favour larger or non-constant settings of the mutation rate, and the optimal mutation rate schedule analysis for a simple objective function provides a good confirmation of the usefulness of larger, varying mutation rates (in classical approaches they are generally $p_m \in [0.001, 0.01]$, see [4,10]). For these reasons, *Bäck* imposed $p_m(0)=½$. However, comparing the GA efficiency based in *Bäck*'s function (Eq. 1 / with $p_m(0)=½$ or $p_m(0)=0.15$ / $T=3000$ / $n=531$), with the *LD/QD* functions, we come up with significant differences (tests #2,3,4,5). These results (although, they are statistically insufficient) probably point that optimal dynamic mutation rates should also be characterised in function of the problem's search landscape (which are manifestly different - "counting ones" *versus* Eq. (1)).

*Bäck* followed this route [3], adapting the mutation rates according to the topology of the objective function, using the principle of strategy parameter self-adaptation as developed by *Schwefel* [20,2] for *Evolution Strategies* (ES), or similarly and independently by *Fogel* [8] for *Evolutionary Programming* (EP, [9]). These models, however, were not applied or have been analysed in the present framework; instead, a novel approach was considered: artificial *neoteny* (*aNeoteny*).

## 3. IMPLEMENTING AND TESTING ARTIFICIAL NEOTENY

In order to implement *aNeoteny*, the preservation of older genotypes is a key-aspect. In general, this preservation was possible through capturing elitist individuals (*neotonic* individuals) from generations $g=0$ till $g=100$ (one per generation), and *throwing* them randomly into later generations (i.e a randomly individual give his place in the population array, to one randomly chosen *neotonic* individual, generally for $g \in [1000,3000]$). Some questions, however, seems to be pertinent. For instance, at which period in the whole evolutionary process should this *neotonic* individuals be captured, how many should be thrown in (in the later generations), and when thus this *throwing* process should start? In order to answer this questions and to evaluate the possible contribution of *Neoteny* in the GA fitness convergence, several tests (38) were conducted (table 1). These tests can be roughly classified into six groups. The first group include tests #1 to #8, and his purpose was to evaluate and compare the GA perfomance for different types of mutation with or without the implementation of neoteny (#7,8) for the same random seed. The second group (tests #10-14) aims to evaluate the same effect but now for a different starting population (the nature of different starting populations can be analysed, in terms of fitness, by table 3). Since the results of the first group suggest the use of neotonic strategies, while the second achieves roughly the same fitness values by incorporating only dynamic mutation rates, the third group (tests #15-25) was dedicated to evaluate if neotonic injection of genotypes could achieve the same results when incorporating that material at different generation intervals (i.e., at different evolutionary periods). Following the same concern, tests #26-30 (fourth group) analyses the effect on the average number of thrown neotonic individuals. The fifth group (tests #31-34) concerned the generation interval where neotonic individuals should be captured, and finally the sixth group (tests #35-38) analyses the effect of re-injecting one neotonic individual simultaneously with one complete random created individual. Average results, for different starting populations and strategies, can be found at table 2, and table 3 presents the random seed effect on some characteristics for these different starting populations used. Finally, fig. 2 presents the convergence of some GA strategies for each generation.

## 4. CONCLUSIONS AND FUTURE WORK

Regarding the neotonic strategies, and by analysing the results of tests #1 to #14 (first and second test groups), it is clear that the strategy of implementing neotonic strategies and dynamic mutation rates can yield substantially (around 58%) the fitness values for the same number of generations (3000), comparing to the use of constant mutation rates (see also table 2). The best result was achieved by using dynamic mutation rates and neotonic strategies (#6), although when we change the starting population the same result was only achieved by using non-neotonic implementations (test #12). It appears that starting populations with above-average individuals on it (see table 3 - random seeds $R$=917, $R$=27 and $R$=7445, which is the case of test group II) do not need for higher exploring natures in the search space to achieve above-average fitness, either by incorporating a slowest decay in the mutation rate (e.g. *LD* versus *QD*) or by yielding the population diversity into the later generations via neotonic strategies. In fact, they appear to achieve good results simple by exploiting the above-average fitness and schema of their population. This is probably why, at constant mutation rates, the starting population with $R$=9 (test #1) with greater average fitness, achieves better results than test #10 ($R$=7445).

It appears also (see tests #15-25) that under these circumstances, no optimal neotonic strategy can be found. In fact, throwing neotonic individuals at different temporal periods point that results can be different and only near fitness values could be found (test #20). However, introducing diversity by neotonic implementations and simultaneously incorporating diversity into this diversity, by adding complete random created individuals (tests #35-38) could yield the fitness values to the same level, for $R$=7445. Apparently this last argument is in contradiction with the one of the last paragraph. However, is the author belief that for some starting populations (e.g., $R$=7445) the increase of diversity (increasing the exploring capabilities of the algorithm) by neotonic strategies cannot fulfil the exploiting power of simple genetic algorithms, unless, this diversity is himself increased. In other words, for a finite number of generations and for the precedent contexts, the best convergence could only be achieved either by increasing the exploring character of the algorithm, or by increasing his exploiting character, that is, renouncing for the suppose-to-be appropriate exploring/exploiting balance. This last point suggests that probably, a diversity *critical-mass* is needed within the evolutionary process, for some starting points in the search landscape.

On the other hand, tests #26-30 suggest that no better results could be found by re-injecting more than one neotonic individual per generation. In fact, results decay has the number of neotonic individuals increases. Finally, results also change if neotonic individuals are captured in different time-windows (tests #31-34 / column F - table 1). Why the interval [1,100] for capturing neotonic individuals, and the interval [1000,3000] for throwing them appear to be optimal, however, is hard to answer. Nevertheless, it appears to be important to give to the evolutionary search some time before re-injecting neotonic individuals, i.e. some evolutionary period where genetic exploitation should be processed in the classical way. Further tests should be implemented in order to analyse this point. Finally, a note about the neotonic strategy effect on the genetic image segmentation processing. In the case of colour images, the differences between both techniques (classical *versus* neotonic) clearly affects the visual quality, namely at enhancing objects extracted (also) by the classical way.

**Acknowledgements**: The author wishes to thank to *David Fogel* (Natural Selection, Inc. / USA), *Thomas Bäck* (Center for Applied Systems Analysis - CASA / Germany) and *Rajeev Ayyagari* (Indian Statistical Institute / India), for their useful references and comments on Dynamic Mutation and *Neoteny* at *COMP.AI.GENETIC* (Jan./Feb. 2000), and also to FCT-PRAXIS XXI (BD20001-99), for his PhD Research Fellowship.